\documentclass[10pt,twocolumn,letterpaper]{article}

\usepackage[final,applications]{wacv}     
\usepackage{multirow}
\usepackage{booktabs}
\usepackage{pifont} 
\usepackage{booktabs}
\usepackage{tabularx}
\usepackage{array}
\usepackage{graphicx}     

\usepackage[utf8]{inputenc} 
\usepackage[T1]{fontenc}    
\usepackage{url}            
\usepackage{booktabs}       
\usepackage{amsfonts}       
\usepackage{nicefrac}       
\usepackage{microtype}      
\usepackage{xcolor}         

\usepackage{docmute}
\usepackage{listings}
\lstdefinestyle{promptstyle}{basicstyle=\ttfamily\scriptsize, breaklines=true, breakatwhitespace=false, columns=fullflexible, keepspaces=true, frame=single, backgroundcolor=\color{gray!8}}

\definecolor{wacvblue}{rgb}{0.21,0.49,0.74}
\usepackage[pagebackref,breaklinks,colorlinks,allcolors=wacvblue]{hyperref}

\def\wacvPaperID{*****} 
\def\confName{WACV}
\def\confYear{2027}

\title{
\raisebox{-0.3\height}{\includegraphics[height=1.5cm]{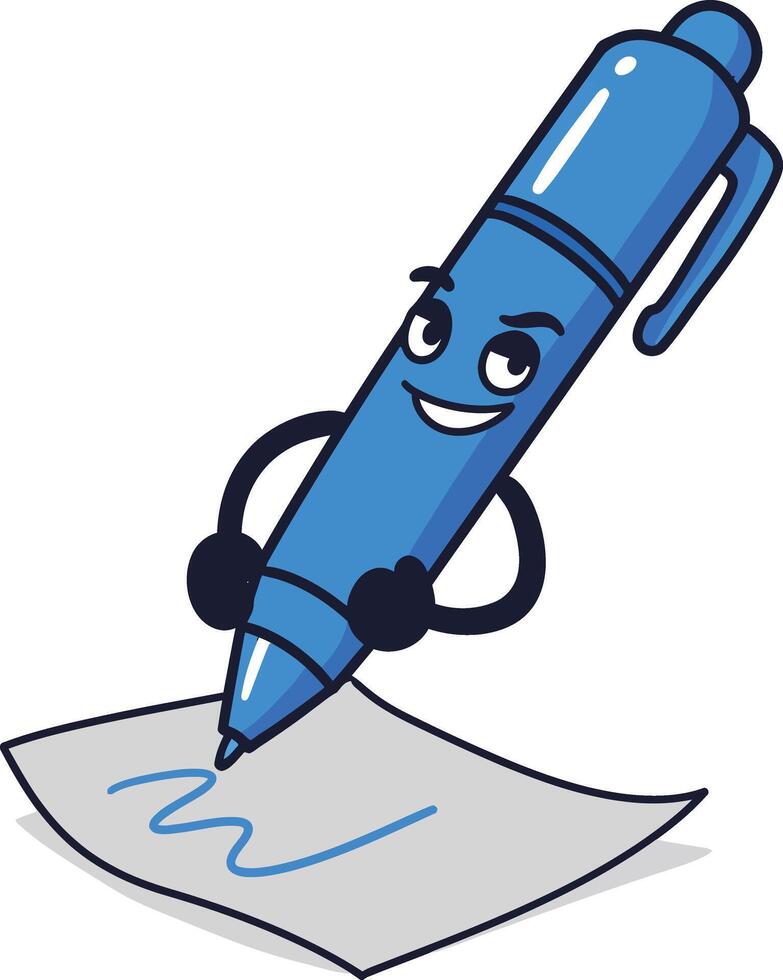}} 
    \hspace{-0.2cm} 
inkGround: Benchmarking Fine-grained Visual Grounding Capabilities of Vision-Language Models for Cursive Handwriting in the Wild
}

\author{Sidra Hnaif\\
San Diego State University\\
{\tt\small shanif@sdsu.edu}}

\begin{document}
\maketitle

\begin{abstract}
Human handwriting is produced as an ordered sequence of pen movements, but an offline image keeps only the final ink. Handwriting trajectory recovery aims to restore the ordered stroke points and their pen-up/pen-down transitions. Existing  image--trajectory pair datasets are recorded with digitizing tablets under controlled conditions, which limits writer coverage and stylistic diversity. Moreover, there is no benchmark to measures the visual grounding capabilities of general-purpose Vision-Language Models (VLMs) for strokes trajectory recovery task. In this work, we introduce \textbf{\textit{inkGround}}, a benchmark for stroke trajectory recovery from full-page, camera-captured handwriting. \textit{\textbf{inkGround}} pairs 537 page images ($\approx$18.7k words, $\approx$163k strokes) with reference trajectories produced by a pretrained offline-to-online model, filtered by OCR agreement, and verified by human raters. Because these references are pseudo-labels that represent one plausible pen trajectory, we also evaluate VLMs on IAM-online datasets, which provides recorded pen trajectories. We cast the task as text-conditioned point grounding: given a page image and a target word, the model must locate the word and output its ordered stroke points. Outputs are scored by trajectory fidelity (DTW, Chamfer distance), readability of the rendered strokes (Accurate Rate, Correct Rate), and style similarity. We benchmark four open-source and three frontier closed-source VLMs with a shared in-context prompt. Closed-source models, notably Gemini-3.1-Pro, produce legible trajectories close to the reference and this ranking also holds for IAM-online datasets.
\end{abstract}    
\section{Introduction}
\begin{figure}[t]
  \centering
  \includegraphics[scale = 0.3]{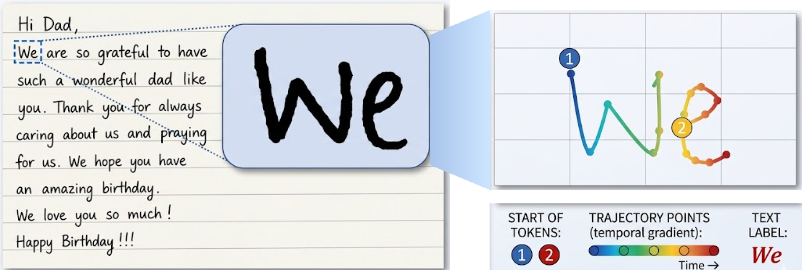}
   \caption{ A sample handwriting image with the word \textit{``We''} highlighted, together with its pen trajectory: start-of-stroke tokens, temporally ordered stroke points (the color gradient indicates writing progression), and the text label.}
   \label{figs:pen_strokes}
\end{figure}
\label{sec:intro}
Handwriting trajectory recovery is a task of restoring the ordered pen movements behind an offline handwriting image, underpins downstream applications such as handwriting biometrics~\cite{recog1}, pen order prediction~\cite{pred,trace,u-trace}, and handwriting style imitation~\cite{vdiff,layout}. Unlike a static image, a stroke trajectory encodes stroke order, pen-up/pen-down transitions and writing dynamics, revealing how characters and words are formed rather than only how they look. Existing recovery systems are specialized, consists of fully supervised networks trained end to end for a single script or dataset~\cite{bhunia2018,trace,u-trace}. Related online generators~\cite{cashg,diffink,vdiff} and character-level methods for Chinese script~\cite{finet,temporal} achieve strong results, but are likewise trained on curated inputs such as isolated characters or clean single-line crops~\cite{casia,IAM-online}.

Meanwhile, Vision-Language Models (VLMs) have matured into general-purpose visual reasoners that follow natural-language instructions~\cite{qwen3,claudeopus,geminipro}. Beyond captioning and question answering, recent VLMs localize objects by returning bounding boxes or points, grounding language directly in the pixel grid. Similarly, document-oriented models~\cite{paddleocr,deepseekocr} read printed and handwritten text straight from images. Yet this progress targets \emph{what} is written, not \emph{how} it was written. We ask whether the grounding and in-context learning abilities of VLMs extend to the stroke level. We cast trajectory recovery as \emph{text-conditioned point grounding}: given a full-page image,  text prompt including an example image with its strokes, the model must locate the word on the page and output its ordered stroke points.
\begin{figure*}[t]
  \centering
  \includegraphics[scale = 0.42]{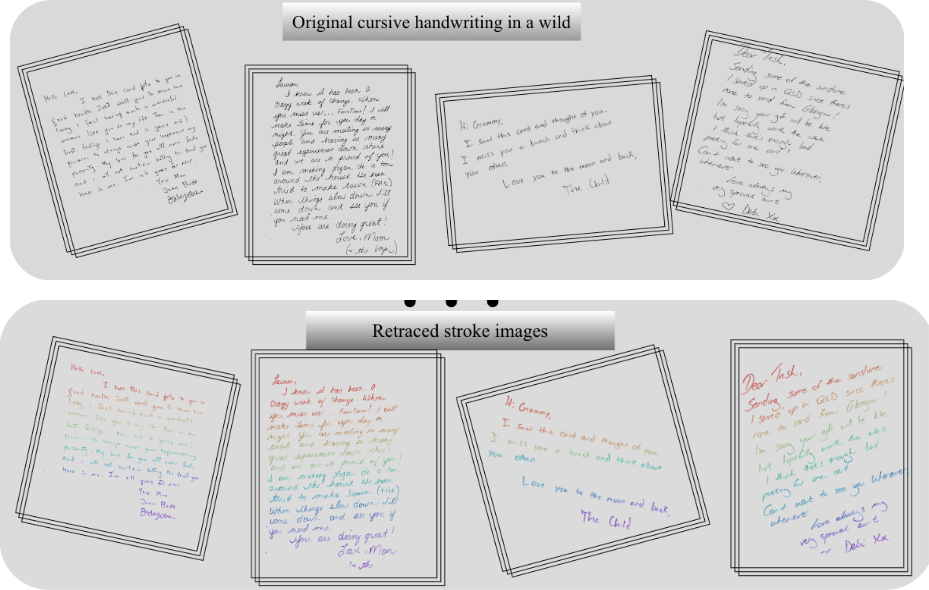}
   \caption{(Sample handwriting images from \textbf{inkGround} and their rendered reference trajectories (best viewed zoomed in).}
   \label{figs:samples}
\end{figure*}

To fully explore the point grounding capabilities of VLMs for stroke trajectory recovery problem, we requires a benchmark that pairs real-world handwriting images in the wild with stroke trajectories. Online datasets recorded with digitizing tablets~\cite{IAM-online,unipen,casia,ub1} provide such pairs, but only for controlled settings.  Mostly as a single-line input from a limited set of writers. To our knowledge, no benchmark evaluates VLM's point grounding capability for stroke recovery on the full-page camera-captured cursive handwriting images.

We address this gap with \textbf{\textit{inkGround}}, a benchmark of 537 camera-captured full-page images with reference stroke trajectories (Fig.~\ref{figs:pen_strokes}). The references are pseudo-labels produced by a pretrained offline-to-online model~\cite{inksight} and retained only after OCR-based filtering and human validation. Because they represent one plausible trajectory rather than recorded pen movements, we complement them with IAM-OnDB~\cite{IAM-online}, which provides recorded trajectories. We evaluate four open-source VLMs (Qwen3-VL-32B~\cite{qwen3}, Molmo-7B-D-0924~\cite{molmo}, InternVL3\_5-8B~\cite{internvl35}, Phi-4-vision-15B~\cite{phi}) and three closed-source VLMs (Gemini-3.1-Pro~\cite{geminipro}, Kimi-K3~\cite{kimik3}, Claude-Opus-4.8~\cite{claudeopus}) with the same prompt and in-context example, and score trajectory agreement, readability of the rendered strokes, and style similarity.
\begin{figure*}[t]
  \centering
  \includegraphics[scale = 0.42]{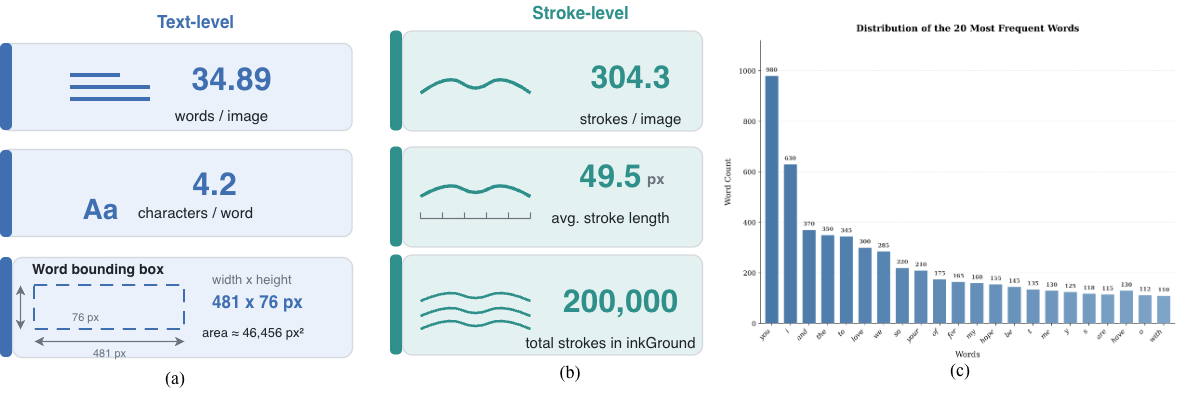}
  \caption{(a) Statistics of the dataset. (b) Distribution of the 20 most frequently occurring words in the benchmark dataset (zoom to view).}
  \label{figs:statistics}
  \vspace{-1em}
\end{figure*}
Our contributions are as follows:
\begin{itemize}
    \item \textbf{Benchmark.} \textit{inkGround} contains 537 full-page, camera-captured handwriting images ($\approx$18.7k words, $\approx$163k strokes) with reference trajectories filtered by OCR agreement and validated by 85 human raters. Every pair is at least rated by two raters. To our knowledge, it is the first full-page image--trajectory benchmark for handwriting captured outside controlled conditions.
    \item \textbf{Task and protocol.} We formulate stroke recovery as text-conditioned point grounding for VLMs and define an evaluation protocol that combines trajectory fidelity (DTW, Chamfer distance), readability (AR, CR), and style similarity.
    \item \textbf{Evaluation.} We benchmark seven VLMs on \textit{inkGround} and on IAM-Online datasets /cite{online}. Closed-source models, led by Gemini-3.1-Pro, produce legible trajectories while open-source models largely fail, the ranking holds on recorded ground truth. Closed-source models also trade readability against style imitation.
\end{itemize}
\section{Related Work}
\label{sec:related_work}
\subsection{Visual grounding in VLMs}
Visual grounding links a phrase or instruction to specific image evidence, such as a bounding box, mask, or point. Segmentation oriented models such as GLaMM~\cite{glamm} and LISA~\cite{lisa} ground language at the pixel level, while Ferret~\cite{ferret} refers to and localizes free-form regions. General-purpose VLMs such as Qwen3-VL~\cite{qwen3} and Molmo~\cite{molmo} output boxes or points directly as text coordinates. MolmoPoint~\cite{molmopoint} argues that textual coordinates are hard to learn and costly in tokens, and instead selects locations from visual features through grounding tokens. RoboPoint~\cite{robopoint} extends point prediction to robotic affordances with a synthetic data pipeline. SketchAgent~\cite{sketchagent} prompts a language model to draw stroke by stroke on a numbered grid, which inspires our prompt format. These methods predict one or a few locations per query, however, stroke recovery requires a long, ordered point sequence that must overlap the ink of a specific word.

\subsection{Handwriting recognition}
Handwritten text recognition has moved from sequence-to-sequence transformers such as TrOCR~\cite{trocr} to document-level systems such as PaddleOCR~\cite{paddleocr} and DeepSeek-OCR~\cite{deepseekocr}, which parse full pages and compress visual context into few tokens. These systems read handwriting reliably, but they recover the transcription rather than the pen trajectory. In this work, We used DeepSeek-OCR to filter our reference trajectories.

\subsection{Stroke trajectory recovery}
Recovering the drawing order of offline handwriting has been approached with recurrent encoder--decoder networks~\cite{bhunia2018,pred}, differentiable line-level rendering~\cite{trace}, and a framework for unconstrained documents with automatic evaluation~\cite{u-trace,hanif2024comprehensive}. InkSight~\cite{inksight} converts offline handwriting into online ink by learning to read and write; we use it to generate our reference trajectories. Related generative work synthesizes online handwriting with context-aware models~\cite{cashg} or latent diffusion~\cite{diffink,vdiff}. \cite{brush} decouples writer style from content for few-shot style imitation and \cite{finet,temporal} reconstructs Chinese character trajectories. These methods are trained per dataset and per script, on full text lines~\cite{trace,u-trace,bhunia2018} or isolated characters~\cite{temporal,finet}, and their performance depends on the scale and diversity of that data. None of these methods evaluates whether a general-purpose VLM can perform the same recovery from in-context examples.

\subsection{Handwriting datasets}
Online datasets such as IAM-online \cite{IAM-online}, UNIPEN~\cite{unipen}, CASIA~\cite{casia}, and IBM\_UB\_1~\cite{ub1} pair handwriting with recorded pen coordinates and pen-up/pen-down events. Since they are collected with digitizing tablets or stylus screens, they cover a limited range of writers and styles and mostly contain isolated characters or single text lines. On the other hand, in-the-wild datasets such as GNHK~\cite{gnhk} provide cluttered, multi-word photographs with varied layouts, but no trajectories. 

\textit{inkGround} combines the two, pairing full-page, camera-captured images with reference trajectories. Some of the examples from \textit{inkGround} are shown in Fig. \ref{figs:samples}. We evaluated VLMs on both \textit{inkGround} and IAM-Online dataset to validate our experiments.

\renewcommand{\dbltopfraction}{0.9}
\begin{figure*}[t]
  \centering
  \includegraphics[width=1\textwidth]{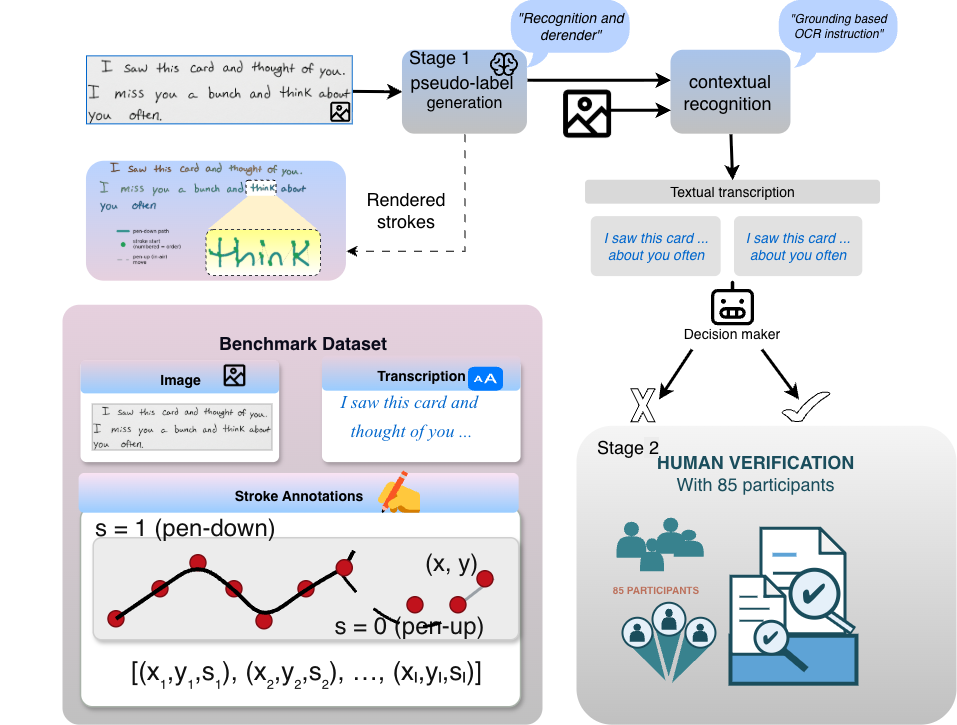}
  \caption{Overview of the benchmark construction pipeline. \textbf{Stage~1:} pseudo-label generation and contextual recognition module. \textbf{Stage~2:} Human validation with 85 participants. \textbf{Benchmark Dataset:} Each benchmark sample consists of original image, the transcription, and stroke annotations encoding sequential coordinates with start-of-stroke tokens.}
  \label{figs:block_diagram}
\end{figure*}
\section{inkGround}
\label{sec:benchmark}
In this section we describe \textbf{inkGround}, a benchmark for stroke trajectory recovery from full-page handwritten images. Sec.~\ref{sec:overview} gives an overview and Sec.~\ref{sec:statistics} reports dataset statistics. The construction pipeline is described in Sec.~\ref{sec:pipeline}.
\subsection{Overview}
\label{sec:overview}
\textbf{inkGround} comprises 537 camera-captured full-page images of handwritten personal notes and greeting-card messages in cursive style. The pages were provided by a greeting-card company [name withheld for review], which holds the copyright and has agreed to the release of the dataset for research purposes. The pages vary in writing style, layout and background. The source provides no writer labels; we therefore treat each page as a separate writer. Each image is paired with its transcription and a reference stroke trajectory $\mathbf{S}=\{(x_t,y_t,s_t)\}_{t=1}^{T}$, where $(x_t,y_t)$ are pen coordinates and $s_t\in\{0,1\}$ is a start-of-stroke token equal to 1 at the first point of each stroke and 0 otherwise. The trajectories are pseudo-labels produced and validated as described in Sec.~\ref{sec:pipeline}. Sample images from \textit{inkGround }are shown in Fig.~\ref{figs:samples} and statistics in Fig.~\ref{figs:statistics}.

\subsection{Statistical Analysis}
\label{sec:statistics}
All statistics in this section are computed on 537 full page of handwritten images.

\noindent
\textbf{Image-level statistics.}
A page contains 34.89 words on average (minimum 2, maximum 140), for $\approx$18.7k words in total (Fig.~\ref{figs:statistics}(a)). Pages therefore range from short notes to dense multi-line messages.

\noindent
\textbf{Bounding box statistics.}
Word-level bounding boxes from the contextual recognition module have a mean (median) width of 0.112 (0.100) of the page width, or 140.7 (124.8) pixels, and a mean (median) height of 0.043 (0.037) of the page height, or 71.9 (63.5) pixels. Individual words thus occupy a small fraction of the page, so a model must localize each word before tracing it.

\noindent
\textbf{Stroke statistics.}
The benchmark contains $\approx$163.4k strokes, with 304.3 strokes per page and 49 points per stroke on average (Fig.~\ref{figs:statistics}(a)).

\noindent
\textbf{Word distribution.}
Fig.~\ref{figs:statistics}(b) shows the 20 most frequent words. Words such as \textit{you}, \textit{I}, \textit{the}, and \textit{and} are the most common, and the high frequency of words such as \textit{love} reflects the personal-note and greeting-card source of the pages.

\section{Benchmark Construction Pipeline}
\label{sec:pipeline}
As offline images contain no temporal information, so we build reference trajectories with the two-stage pipeline in Fig.~\ref{figs:block_diagram}. In Stage~1 (Sec.~\ref{sec:inksight}), a pretrained offline-to-online model generates pseudo-label trajectories, which are rendered as images and filtered by OCR agreement. In Stage~2 (Sec.~\ref{sec:human}), human raters validate the remaining image--trajectory pairs to make sure that all the reference trajectories overlaps with the handwriting ink.

\subsection{Stage 1: Pseudo-label generation and filtering}
\label{sec:inksight}
We use the publicly released InkSight Small-p model~\cite{inksight}, which predicts stroke tokens from a handwriting image with the prompt \textit{"Recognize and derender."} InkSight's built-in text detection segments each full page into words; every word is derendered, and the resulting strokes are placed back in page coordinates. The output is a sequence of $(x_t,y_t,s_t)$ tuples. When we render the strokes with a color gradient, it shows the temporal order of writing as shown in Fig.~\ref{figs:pen_strokes}. On camera-captured pages with complex layouts and overlapping strokes, the model predictions can miss strokes or produce misaligned strokes, so we filter them as described below. 
\begin{table*}[t]
\centering
\tiny
\resizebox{\textwidth}{!}{%
\begin{tabular}{l ccccc}
\toprule
\textbf{Method} & \textbf{DTW} $\downarrow$ & \textbf{Chamfer Distance} $\downarrow$ & \textbf{AR} $\uparrow$ & \textbf{CR} $\uparrow$ & \textbf{Style} $\uparrow$ \\
\midrule
\multicolumn{6}{c}{\textit{Open-source VLMs}} \\
\midrule
Qwen3-VL-32B~\cite{qwen3} & 0.378 & 0.222 & 0.0529 & 0.0529 & 0.0909 \\
Molmo-7B-D-0924~\cite{molmo} & 0.463 & 0.650 & 0.0754 & 0.0820 & 0.0965 \\
InternVL3\_5-8B~\cite{internvl35} & 0.288 & 0.533 & 0.0344 & 0.0344 & 0.0754 \\
Phi-4-vision-15B~\cite{phi} & 0.389 & 0.639 & 0.0696 & 0.0699 & 0.0703 \\
\midrule
\multicolumn{6}{c}{\textit{Frontier proprietary VLMs}} \\
\midrule
Claude-Opus-4.8~\cite{claudeopus} & 0.209 & 0.119 & 0.768 & 0.812 & 0.283 \\
Gemini-3.1-Pro~\cite{geminipro} & \textbf{0.105} & \textbf{0.052} & \textbf{0.812} & \textbf{0.826} & 0.861 \\
Kimi-K3~\cite{kimik3} & 0.171 & 0.091 & 0.581 & 0.591 & \textbf{0.927} \\
\bottomrule
\end{tabular}
}
\caption{Evaluation of open and frontier closed-source VLMs on \textbf{inkGround} under the shared prompting protocol. DTW and Chamfer distance are computed against the reference pseudo-label trajectories; AR and CR measure the readability of the rendered strokes; Style is the writer-classifier score with each page treated as one writer (Sec.~\ref{sec:metrics}). Invalid or incomplete outputs are scored as described in Sec.~\ref{sec:failures}.}
\label{tab:inkground}
\end{table*}
\vspace{3pt}
\noindent
\textbf{Contextual recognition.}
We use DeepSeek-OCR~\cite{deepseekocr}, a VLM-based OCR system that reads the full page and uses linguistic context across lines, unlike OCR systems that process words or lines in isolation~\cite{trocr,paddleocr}. The grounding prompt \textit{``$<$image$>$ $<|$grounding$|>$ OCR this image.''} returns both the transcription and word-level bounding boxes. Decoding details are given in the supplementary material (Appendix B).

\vspace{-1pt}
\noindent
\textbf{Filtering.}
If a trajectory preserves the content of the page, its rendering should read the same as the original. We transcribe both the original image and the rendered trajectory with the contextual OCR and compute a word-level matching score, the fraction of words that match between the two transcriptions. Pages with a score below 0.85, i.e., with fewer than 85\% matching words, are discarded. The threshold was chosen empirically to balance pseudo-label quality against dataset size. Of 2,984 candidate pages, 777 (26\%) pass then 85\% matching words threshold. The OCR agreement confirms the text but not the appearance of the writing, so every retained pair is also validated by humans.

\vspace{-3pt}
\noindent
\subsection{Stage 2: Human validation}
\label{sec:human}
We presented 777 pages filtered by the contextual OCR to 85 participants (aged 18--50) for human validation. Each participant sees an original page next to the rendering of its Stage~1 trajectory and rates their similarity from 0 (completely different) to 10 (identical, with no missing trajectories or misalignment). Starting from 10, raters deduct one point for each missing word and each word whose writing style differs from the original. The full set of instructions given to the participants are given in Appendix~C of the supplementary material. Each pair is rated independently by at least two participants and is accepted only if every rater gives a score of 10. The raters agree on the accept/reject decision for 81.7\% of pairs. After validation, 537 of the 777 pairs are retained. All statistics in Sec.~\ref{sec:statistics} and all results in Table \ref{tab:inkground} are computed on this final set.

Each \textbf{\textit{inkGround}} sample (Fig.~\ref{figs:block_diagram}, bottom left) consists of (i)~the original camera-captured page image; (ii)~its transcription and (iii)~the reference stroke trajectory $\mathbf{S}=\{(x_t,y_t,s_t)\}_{t=1}^{T}$.

Even after filtering, a trajectory is confirmed only to reproduce the text and appearance of the image, not the writer's original stroke order or direction, since several trajectories can render to a similar image. We therefore treat the retained trajectories as \textit{reference pseudo-labels} that represent one plausible writing order, not as recorded ground truth.

\section{Experiments}
\label{sec:experiments}
We evaluate VLMs on \textit{\textbf{inkGround}} (pseudo-label references) and on IAM-online~\cite{IAM-online} (recorded ground-truths). In both cases, a model receives the full page image and an in-context example, and must output the ordered stroke points of that word. We score the outputs with metrics for trajectory fidelity, readability, and style.
\subsection{Evaluation Metrics}
\label{sec:metrics}

\vspace{3pt}
\noindent
\textbf{Dynamic Time Warping (DTW)} measures sequence-level similarity by optimally aligning stroke trajectories of varying lengths, making it well suited for comparing temporal stroke dynamics ~\cite{u-trace}. 

\vspace{3pt} \noindent \textbf{Chamfer Distance} quantifies the geometric similarity between generated and ground-truth stroke point sets, measuring the spatial alignment of stroke trajectories. Both DTW and Chamfer distance~\cite{hanif2024comprehensive} are computed per word on coordinates normalized by the word's height and averaged over all words on a page, so we desire to have a lower values for both DTW and chamfer distance. 

\begin{table*}[t]
\centering
\tiny
\resizebox{\textwidth}{!}{%
\begin{tabular}{l ccccc}
\toprule
\textbf{Method}
& \textbf{DTW} $\downarrow$
& \textbf{Chamfer Distance} $\downarrow$
& \textbf{AR} $\uparrow$
& \textbf{CR} $\uparrow$
& \textbf{Style} $\uparrow$ \\
\midrule
\multicolumn{6}{c}{\textit{Domain model}} \\
\midrule
inksight ~\cite{inksight}      & 0.062 & 0.042 & 0.9077 &  0.8692 & 0.9667\\
\midrule
\multicolumn{6}{c}{\textit{Open-source} VLMs} \\
\midrule
Qwen3-VL-32B~\cite{qwen3}      & 1.518 & 6.09 & 0.0519  & 0.0519 & 0.6179 \\
Molmo-7B-D-0924~\cite{molmo}         & 7.743 & 4.897 & 0.0047 & 0.0047 & 0.6368 \\
InternVL3\_5-8B~\cite{internvl35}      & 4.729 & 2.487 & 0.0236 & 0.0236& 0.5094 \\
Phi-4-vision-15B~\cite{phi}         & 4.742 & 3.250 & 0.0991& 0.0991 & 0.6415 \\
\midrule
\multicolumn{6}{c}{\textit{Frontier proprietary VLMs}} \\
\midrule
Claude-Opus-4.8~\cite{claudeopus}        & 0.1365 & 0.0670 & 0.8208	& 0.8208 & 0.6667  \\
Gemini-3.1-Pro~\cite{geminipro}        & \textbf{0.088} & \textbf{0.047} & \textbf{0.8644} & \textbf{0.8562} & 0.8889 \\
Kimi-K3~\cite{kimik3}       & 0.158 & 0.087 & 0.4811 & 0.4858 & \textbf{0.8951}\\
\bottomrule
\end{tabular}
}
\caption{Evaluation on IAM-online dataset, which provides recorded pen trajectories. DTW and Chamfer distance are computed against the recorded ground-truth trajectories; AR and CR measure the readability of the rendered strokes; Style is the writer-classifier score over the 217 IAM-OnDB writers (Sec.~\ref{sec:metrics}). InkSight is a specialist offline-to-online model trained on the IAM-OnDB training set and evaluated on the same test set; it is included as an in-domain reference. Invalid or incomplete outputs are scored as described in Sec.~\ref{sec:failures}.}
\label{tab:iam}
\end{table*}

\vspace{3pt} \noindent \textbf{Accurate Rate (AR) and Correct Rate (CR).} To evaluate readability, we render the predicted strokes as an image, transcribe it with Deepseek-OCR, and compare the result with the reference transcription~\cite{ar}. With $N$ reference characters and $S$, $D$ and $I$ substitution, deletion and insertion errors, $\mathrm{AR} = (N - S - D - I)/N$ and $\mathrm{CR} = (N - S - D)/N$. AR therefore additionally penalizes inserted characters.
\vspace{3pt} \noindent \textbf{Style.} To evaluate visual fidelity, we measure how similar the writing style of the rendered strokes is to that of the original image with a writer-identification classifier. The classifier is an ImageNet-pretrained ResNet-18 fine-tuned on  word crops. For IAM-OnDB (Table~\ref{tab:iam}), it is trained on the recorded trajectories rendered as images for 217 writers in the dataset. \textit{inkGround} has no writer labels, so for Table~\ref{tab:inkground} each of the 537 pages is treated as one writer and the classifier is trained on word crops rendered from the reference trajectories, with 75\% of each page's words used for training and 25\% held out for testing. The style score of a rendered word is the probability that the classifier assigns to its true writer, averaged over all evaluated words. Because the IAM-OnDB classifier chooses among 217 writers and the \textit{inkGround} classifier among 537 pages, the two settings have different chance levels and score ranges, so style scores are comparable within a table but not across Tables~\ref{tab:iam} and~\ref{tab:inkground}.
\begin{figure*}[!t]
  \centering
  \includegraphics[scale=0.3]{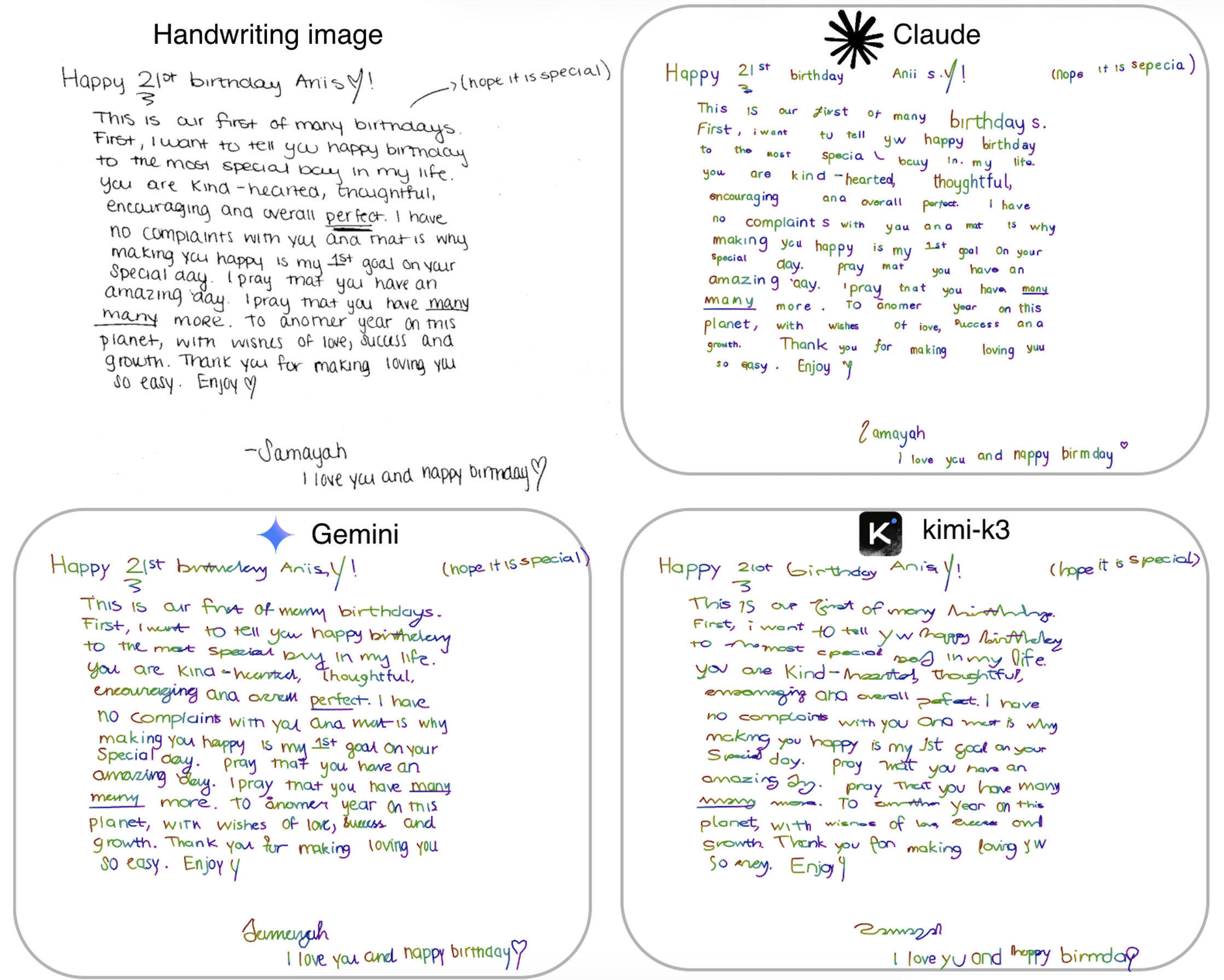}\hfill
  \includegraphics[scale =0.3]{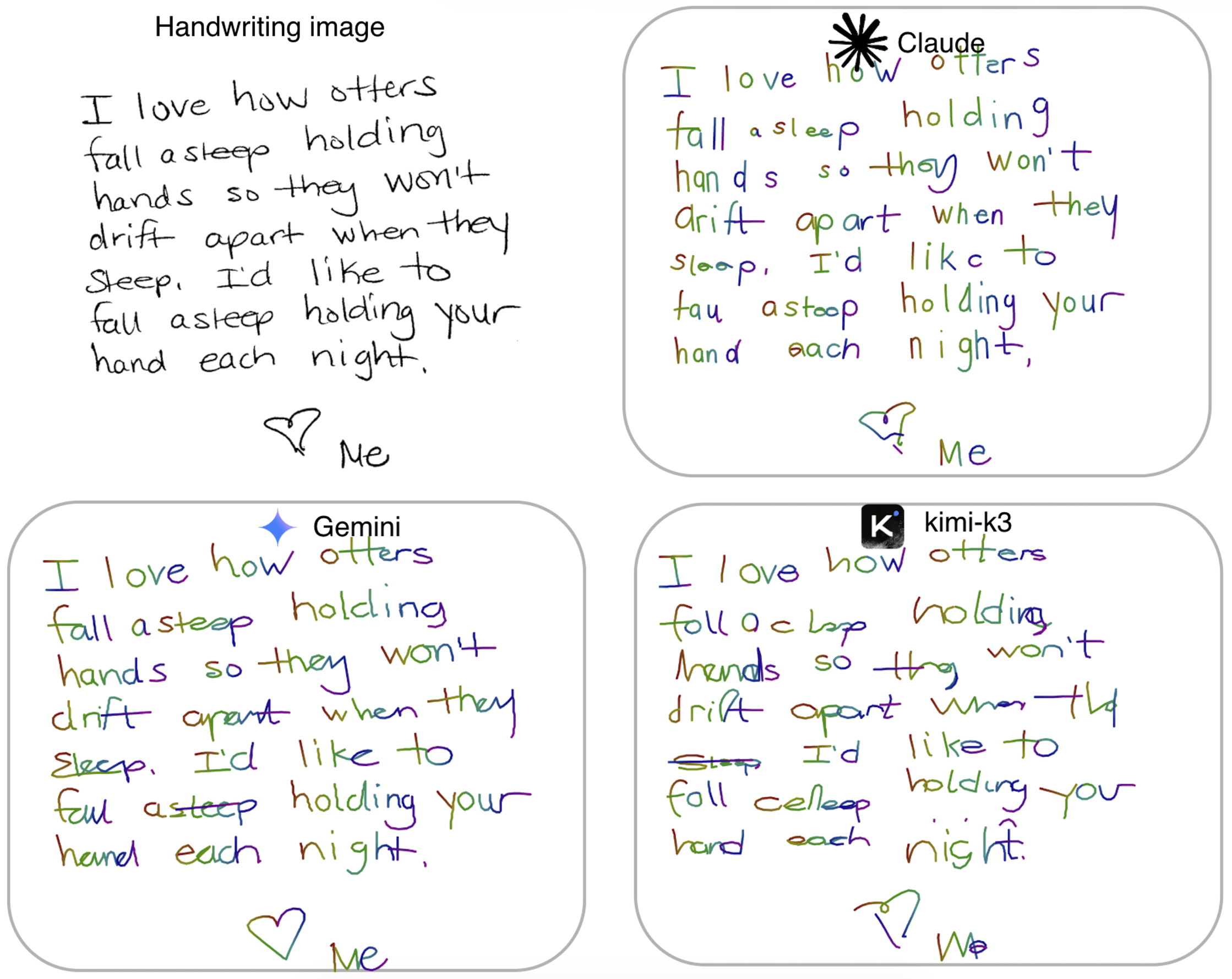}
   \caption{Qualitative results with Claude-Opus-4.8, Gemini-3.1-Pro and Kimi-K3 on two \textit{inkGround} examples (top and bottom). In each example, the input handwriting image is shown with the strokes rendered from each model.}
   \label{figs:qualitative}
\end{figure*}
\subsection{Inference protocol and output validity}
\label{sec:failures}
All models receive the same system prompt, task prompt and in-context example (as given in supplementary material, Appendix A). An output is \emph{invalid} if it cannot be parsed into the required stroke format, and \emph{incomplete} if it covers only part of the words on the page. Invalid outputs receive AR = CR = Style = 0 and the maximum DTW and Chamfer distance. 

On \textit{inkGround}, all models are evaluated on the 537 pages. On IAM-OnDB, all models are evaluated on the test set; input images are rendered from the recorded online strokes, and the in-context example uses the same format with an IAM-online word from a different writer than the test sample.

\subsection{Model evaluation}
\label{sec:baseline_analysis}
We evaluate the point-grounding ability of four open-source and three frontier closed-source VLMs on stroke trajectory recovery. The results are summarized in Tables~\ref{tab:iam} and~\ref{tab:inkground}.
\vspace{3pt}
\noindent
\textbf{Open-source VLMs:}
We select four open-source VLMs namely Qwen3-VL-32B \cite{qwen3}, Molmo-7B-D-0924~\cite{molmo}, InternVL3\_5-8B~\cite{internvl35} and Phi-4-vision-15B~\cite{phi}. Each model receives the system prompt, the task prompt and an example image--stroke pair that shows the required output format. All the tested open-source models struggle to produce readable strokes. Even Qwen3-VL-32B and Molmo-7B-D-0924, which have point-grounding ability, struggle with the stroke trajectory recovery task. Although they attempt to follow the trajectory of the word but do not produce readable text. On \textit{inkGround} (Table~\ref{tab:inkground}), their DTW and Chamfer distance are higher than those of the closed-source models, and their AR and CR stay below 0.08, so their rendered strokes are not readable. Their style scores are also low. Therefore, under our prompting setup, the tested open-source models which do not reliably perform the point-grounding for ink rendering task. 

\vspace{3pt}
\noindent
\textbf{Closed-source VLMs:}
For the frontier closed source model experimentation, the same prompt is provided to Claude-Opus-4.8~\cite{claudeopus}, Gemini-3.1-Pro~\cite{geminipro} and Kimi-K3~\cite{kimik3}. 
The closed-source models produce readable output, as indicated by their high AR and CR metrics. Among them, Kimi-K3 has the lowest readability with 0.581(\textit{AR}) and 0.591(\textit{CR}). Claude-Opus-4.8 has reasonable readability, and Gemini-3.1-Pro shows the highest values of 0.812(\textit{AR}) and 0.826(\textit{CR}). Although Kimi-K3 has the lowest AR and CR, it achieves the highest style score (Table~\ref{tab:inkground}). In contrast, Claude-Opus-4.8 shows the weakest style imitation. Claude-Opus-4.8 produces readable text but does not imitate the style of the handwriting image. The qualitative results for three closed-source model are shown in Fig.~\ref{figs:qualitative}. The examples also show that Kimi-K3 imitates style best but struggles to generate readable text, Claude-Opus-4.8 produces readable text but compromises on style imitation, and Gemini-3.1-Pro balances the two metrics. More examples are given in the supplementary material (Appendix D).

\vspace{3pt}
\noindent
\textbf{Recorded ground truth (IAM-online dataset).}
Since the \textit{inkGround} references are pseudo-labels from a pretrained model, we also evaluate on IAM-online datasets, whose ground-truth trajectories were recorded during writing.
Table~\ref{tab:iam} evaluates the pretrained models \cite{inksight} against recorded pen trajectories constructed by human annotations. The specialist InkSight model provides a reference point (DTW 0.062, Chamfer 0.042). Gemini-3.1-Pro comes closest to it (DTW 0.088, Chamfer 0.047, AR 0.856), followed by Claude-Opus-4.8 (DTW 0.137, AR 0.821), while the open-source models remain far behind, with AR below 0.10 and DTW above 1.5. Kimi-K3 again has the lowest readability among the closed-source models (AR 0.486) but the highest style score (0.895), repeating the readability--style trade-off observed on \textit{inkGround}. InkSight is trained on the IAM-online training set, so its scores serve as an in-domain reference rather than a direct comparison. The ranking Gemini-3.1-Pro $>$ Claude-Opus-4.8 $>$ open-source models is the same as on \textit{inkGround}, which indicates that this comparison is not an artifact of the pseudo-label references. 


\section{Discussion}

All closed-source VLMs recover readable trajectories, but they trade style imitation against readability. Claude-Opus-4.8 recovers strokes of cursive handwriting effectively but ignores writer styles. Kimi-K3 favors style imitation, but its text is harder to read than that of Claude-Opus-4.8 and Gemini-3.1-Pro. Gemini-3.1-Pro shows satisfactory style imitation and reasonable readability, but there is still room for improvement in AR (0.812) and CR (0.826). These results underscore that stroke trajectory recovery from full-page handwriting remains an open challenge, with no tested VLM addressing all facets of the task. In our experimentation, all the open-source models are run on NVIDIA A100 GPUs.

\vspace{3pt}
\noindent
\textbf{Limitations.} The reference trajectories are pseudo-labels generated by a pretrained model. OCR agreement and human validation confirm that they reproduce the text and appearance of the image, but they cannot confirm the writer's original stroke order or direction, and several trajectories may render to a similar image. Consequently, DTW and Chamfer distance measure agreement with a plausible reference rather than recovery of the actual pen movements, and DTW may penalize valid alternative orders. The evaluation on IAM-online with recorded trajectories (Table~\ref{tab:iam}) addresses this concern, since it yields the same model ranking as \textit{inkGround}. We recommend human validation before using the trajectories as training targets. The pseudo-labels come from a small public model, and a larger model may yield better references. The same OCR model (DeepSeek-OCR) is used both to filter the references and to compute AR and CR, which may favor outputs that resemble the references, an independent OCR model would give a stricter readability measure. Finally, our comparison is specific to the tested model versions and to a single shared prompt with in-context examples; other prompting strategies, fine-tuning, or tool use may change the ranking.


\section{Conclusion} \label{sec:conclusion}
We propose \textbf{inkGround}, a benchmark that pairs camera-captured, full-page handwriting images with reference stroke trajectories, and we use it to evaluate VLMs on stroke trajectory recovery. Existing handwriting datasets either lack stroke-level supervision or focus on isolated characters, making them insufficient for benchmarking  handwriting in the wild. Under a shared in-context prompting protocol, we tested closed-source VLMs recover legible trajectories while the open-source models are not able to recover stroke trajectory from handwriting images. The same ranking holds on IAM-online with recorded ground truth.
{
    \small
    \bibliographystyle{ieeenat_fullname}
    \bibliography{main}
}

\clearpage
\appendix

\section{Supplementary material}
This supplementary material provides (A) the exact prompts and the in-context example used to query all VLMs, (B) details of the contextual recognition module used for filtering, (C) the instructions given to the human raters, (D) additional qualitative results from the closed-source models, and (E) the data source.


\subsection*{Appendix A: Prompts for Stroke Trajectory Recovery}
\label{app:prompts}

We frame handwriting stroke trajectory recovery as the task of recovering the
pen strokes from a handwriting image: when the recovered strokes are rendered
back onto the image, they should overlap the handwritten text and follow its
trajectory. Each stroke is a sequence of $(x, y)$ pen coordinates paired with a
start-of-stroke (SOS) token, where the token is $1$ at the point where a stroke
begins (i.e., where the pen is lifted and set down again) and $0$ for every
continuation point. The input is a handwriting image that may contain multiple
words; the VLM first detects the words and then predicts the stroke trajectory
for each word. Following the prompting format of SketchAgent (Vinker et al., 2024), we provide the model with
sample handwriting images together with their ground-truth strokes as in-context
examples. The exact system prompt, task prompt, and a worked ground-truth
example (for the word \emph{forever}) that we use to query the model are given
below. The same system prompt, task prompt and output format are used for IAM-OnDB, with the in-context example taken from an IAM-OnDB word written by a different writer than the test sample.

\paragraph{System prompt.}
\begin{lstlisting}[style=promptstyle]
You are an expert in handwriting analysis, specialized in tracing the pen
trajectory of the handwriting in a given image. The predicted trajectory should
result in readable handwritten text. When text recognition is applied to the
input handwriting image and to the image rendered from the predicted stroke
trajectory, both should produce the same recognition result.

The input image is overlaid with a coordinate grid. {res_x} and {res_y} are the
maximum x and maximum y values of the grid. The grid uses numbers (1 to {res_x})
along the top edge (x axis) and numbers (1 to {res_y}) along the left edge
(y axis). The origin is at the top-left corner: x increases to the right and
y increases downward. Each cell is uniquely identified by its x and y numbers
(e.g., the top-left cell is 'x1y1', the cell to its right is 'x2y1', and the
cell below it is 'x1y2'). Your task is to specify strokes on this grid that
trace the handwritten word in the image, with the same writing style and the
same recognition result. A word may require multiple strokes; different
strokes compose different parts of the word.

To draw a stroke on the grid, specify the following:
Start-of-stroke token: Each output point is 3-dimensional; the first number is
the x-coordinate, the second number is the y-coordinate, and the third number
is the start-of-stroke token. The start-of-stroke token is 1 if the point is
the first point of a stroke (the pen is put down) and 0 otherwise. For example,
if a stroke starts at cell x1y1 (column 1, row 1), its first point is (1, 1, 1).
Intermediate points: Specify at least two intermediate points that the stroke
passes through, listed in the order the pen follows them, using the same grid
format (e.g., 'x6y5', 'x13y10' for points at column 6 row 5 and column 13
row 10).

Examples:
To draw a smooth curve that starts at x8y6, passes through x6y7 and x6y10, and
ends at x8y11:
  Points = ['x8y6', 'x6y7', 'x6y10', 'x8y11']
  values = [(8, 6, 1), (6, 7, 0), (6, 10, 0), (8, 11, 0)]
To close this curve into an ellipse, add another curve:
  Points = ['x8y11', 'x11y10', 'x11y7', 'x8y6']
  values = [(8, 11, 1), (11, 10, 0), (11, 7, 0), (8, 6, 0)]
To draw shapes with sharp corners, repeat each corner point twice. For example,
a "V" shape that starts at x13y27, ends at x24y27, with its lowest point
(corner) at x18y37:
  Points = ['x13y27', 'x18y37', 'x18y37', 'x24y27']
  values = [(13, 27, 1), (18, 37, 0), (18, 37, 0), (24, 27, 0)]
To draw a single dot (e.g., the dot of an 'i') at x15y31:
  Points = ['x15y31']
  values = [(15, 31, 1)]
To draw a straight line (e.g., the crossbar of a 't') from x18y31 to x35y31:
  Points = ['x18y31', 'x35y31']
  values = [(18, 31, 1), (35, 31, 0)]

The points of a stroke are connected in the listed order to form the rendered
trajectory. For a long stroke, use enough points to follow its curvature.
To draw correct and readable text, break complex strokes into manageable steps.
Draw the strokes in the order a writer would produce them, generally from left
to right. When the pen is lifted, finish the current stroke and begin a new one
with a start-of-stroke token of 1. After drawing all strokes, make sure that
all parts of the word are covered. The strokes should result in readable text
with the same style as the input image. Think step by step.
\end{lstlisting}

\paragraph{Task prompt.}
\begin{lstlisting}[style=promptstyle]
I provide you with an input image. Your goal is to produce a correct stroke
trajectory for the word {concept}. Here are a few examples:
<examples> {gt_sketches_str} </examples>
Provide x-y coordinates that construct a stroke trajectory for the word
{concept} in the handwriting image. When the trajectory is plotted, it should
reconstruct the word {concept}. Note that you will not have access to any
additional resources. Do not copy the stroke trajectory of previous words.

Think before you provide the x-y coordinates, inside <thinking> tags. First,
determine where the first stroke of the word {concept} starts. The word may
contain multiple strokes; determine the starting point of every stroke and
their order. Then determine where each start-of-stroke point and the
trajectory that follows it lie on the grid. Finally, provide your response
inside <answer> tags, in the following format:
<formatting>
<concept>The text in the handwriting image.</concept>
<strokes>
  This element holds the individual strokes that reconstruct the handwriting
  in the image. Each stroke is uniquely identified by its own tag (e.g., <s1>,
  <s2>, etc.). Each stroke element contains three fields:
  <id>A short descriptive identifier explaining which part of the word the
    stroke corresponds to.</id>
  <points>The ordered list of grid cells the stroke passes through
    (e.g., x6y5, x13y10).</points>
  <values>The same points as (x, y, sos) tuples, where sos is the
    start-of-stroke token: 1 for the first point of the stroke (the pen is
    put down) and 0 for every following point.</values>
</strokes>
</formatting>
\end{lstlisting}

\paragraph{Ground-truth example (word \emph{forever}).}
\begin{lstlisting}[style=promptstyle]
<example>
To draw the word 'forever', we draw it stroke by stroke:
<concept>forever</concept>
<strokes>

<s1>
  <id>f body -- curved descending stroke with top arc</id>
  <points>
    x59y95, x59y93, x59y91, x58y88, x57y86, x55y85, x53y84, x51y84, x49y85,
    x47y87, x45y90, x44y94, x43y99, x43y105, x43y111, x44y118, x45y124, x46y129,
    x47y133, x48y136, x49y138
  </points>
  <values>
    (59,95,1), (59,93,0), (59,91,0), (58,88,0), (57,86,0), (55,85,0), (53,84,0),
    (51,84,0), (49,85,0), (47,87,0), (45,90,0), (44,94,0), (43,99,0), (43,105,0),
    (43,111,0), (44,118,0), (45,124,0), (46,129,0), (47,133,0), (48,136,0),
    (49,138,0)
  </values>
</s1>

<s2>
  <id>f crossbar -- short horizontal line</id>
  <points>
    x34y116, x35y116, x37y116, x40y116, x43y115, x47y114, x51y113
  </points>
  <values>
    (34,116,1), (35,116,0), (37,116,0), (40,116,0), (43,115,0), (47,114,0),
    (51,113,0)
  </values>
</s2>

<s3>
  <id>o -- closed oval loop</id>
  <points>
    x73y109, x72y108, x71y107, x69y107, x67y107, x65y108, x63y109, x61y111,
    x59y114, x58y117, x58y121, x59y125, x61y128, x63y130, x66y131, x69y131,
    x72y130, x75y128, x77y125, x78y122, x78y118, x78y115, x77y112, x75y110,
    x73y108, x71y107, x69y107
  </points>
  <values>
    (73,109,1), (72,108,0), (71,107,0), (69,107,0), (67,107,0), (65,108,0),
    (63,109,0), (61,111,0), (59,114,0), (58,117,0), (58,121,0), (59,125,0),
    (61,128,0), (63,130,0), (66,131,0), (69,131,0), (72,130,0), (75,128,0),
    (77,125,0), (78,122,0), (78,118,0), (78,115,0), (77,112,0), (75,110,0),
    (73,108,0), (71,107,0), (69,107,0)
  </values>
</s3>

<s4>
  <id>r stem -- short vertical ascending stroke</id>
  <points>
    x90y132, x90y131, x90y129, x90y126, x90y123, x90y119, x90y115, x90y111,
    x91y108, x91y105, x92y103, x93y102
  </points>
  <values>
    (90,132,1), (90,131,0), (90,129,0), (90,126,0), (90,123,0), (90,119,0),
    (90,115,0), (90,111,0), (91,108,0), (91,105,0), (92,103,0), (93,102,0)
  </values>
</s4>

<s5>
  <id>e -- open oval loop with flat top</id>
  <points>
    x102y122, x103y123, x105y123, x107y123, x109y122, x111y121, x113y119,
    x114y117, x114y115, x114y113, x113y111, x112y110, x110y109, x108y109,
    x106y109, x104y110, x103y112, x102y115, x101y118, x101y122, x102y126,
    x103y129, x105y132, x107y134, x110y135, x113y135, x116y134, x118y132
  </points>
  <values>
    (102,122,1), (103,123,0), (105,123,0), (107,123,0), (109,122,0), (111,121,0),
    (113,119,0), (114,117,0), (114,115,0), (114,113,0), (113,111,0), (112,110,0),
    (110,109,0), (108,109,0), (106,109,0), (104,110,0), (103,112,0), (102,115,0),
    (101,118,0), (101,122,0), (102,126,0), (103,129,0), (105,132,0), (107,134,0),
    (110,135,0), (113,135,0), (116,134,0), (118,132,0)
  </values>
</s5>

<s6>
  <id>v -- V-shaped curve descending then ascending</id>
  <points>
    x124y109, x124y110, x124y112, x125y115, x126y122, x127y126, x128y129,
    x129y132, x130y134, x131y135, x132y135, x133y134, x134y132, x135y129,
    x136y125, x136y121, x136y117, x136y113, x136y110, x136y108, x136y107
  </points>
  <values>
    (124,109,1), (124,110,0), (124,112,0), (125,115,0), (126,122,0), (127,126,0),
    (128,129,0), (129,132,0), (130,134,0), (131,135,0), (132,135,0), (133,134,0),
    (134,132,0), (135,129,0), (136,125,0), (136,121,0), (136,117,0), (136,113,0),
    (136,110,0), (136,108,0), (136,107,0)
  </values>
</s6>

<s7>
  <id>er -- partial oval loop for e then extending tail curving right for r</id>
  <points>
    x145y122, x146y122, x147y122, x149y122, x151y121, x153y120, x155y118,
    x156y116, x156y114, x156y113, x155y112, x153y111, x151y111, x149y112,
    x147y113, x146y115, x145y118, x145y121, x146y124, x147y127, x149y130,
    x152y132, x155y133, x158y133, x161y132, x163y130, x165y128, x166y125,
    x167y122, x167y119, x167y117, x167y115, x167y114, x167y113, x168y112,
    x169y111, x171y110, x173y109, x175y108, x178y107, x181y106, x184y106,
    x187y106, x190y107
  </points>
  <values>
    (145,122,1), (146,122,0), (147,122,0), (149,122,0), (151,121,0), (153,120,0),
    (155,118,0), (156,116,0), (156,114,0), (156,113,0), (155,112,0), (153,111,0),
    (151,111,0), (149,112,0), (147,113,0), (146,115,0), (145,118,0), (145,121,0),
    (146,124,0), (147,127,0), (149,130,0), (152,132,0), (155,133,0), (158,133,0),
    (161,132,0), (163,130,0), (165,128,0), (166,125,0), (167,122,0), (167,119,0),
    (167,117,0), (167,115,0), (167,114,0), (167,113,0), (168,112,0), (169,111,0),
    (171,110,0), (173,109,0), (175,108,0), (178,107,0), (181,106,0), (184,106,0),
    (187,106,0), (190,107,0)
  </values>
</s7>

</strokes>
</example>
\end{lstlisting}

\subsection*{Appendix B: Contextual recognition module}

Given an input image $\mathbf{I}$, the contextual OCR first compresses the high-dimensional visual input into compact contextual tokens using a deep encoder:
\begin{equation}
\mathbf{Z} = \mathrm{Enc}_{v}(\mathbf{I}),
\end{equation}
where $\mathbf{Z}$ captures glyph structure, spatial layout, and visual context while discarding redundant visual details. Rather than performing explicit character-level segmentation, the model treats these embeddings as a compressed optical representation that is interpreted through a language-aware decoder. This decoder, conditioned on the visual tokens, autoregressively generates the output text sequence $\mathbf{Y} = \{y_1, y_2, \ldots, y_T\}$:
\begin{equation}
p(\mathbf{Y} \mid \mathbf{I}) = \prod_{t=1}^{T} p(y_t \mid \mathbf{Z}, y_{<t}).
\end{equation}

\subsection*{Appendix C: Instructions for human evaluation}
\label{app:human}
Each participant compares an original page with the rendering of its Stage~1 trajectory and starts from a score of 10. The participants are instructed to:
\begin{itemize}
\item read the original and the rendered image and match each word between the two;
\item check that the writing style of the rendered image matches the original;
\item deduct one point for each character or spelling error in the rendered image;
\item deduct one point for each word whose style is inconsistent with the original;
\item deduct one point for each word missing from the rendered image;
\item assign a score of zero if the rendered image contains more than 10 errors.
\end{itemize}

\subsection*{Appendix D: Additional qualitative results}
\begin{figure*}[p]
  \centering
  \includegraphics[width=\textwidth,height=0.42\textheight,keepaspectratio]{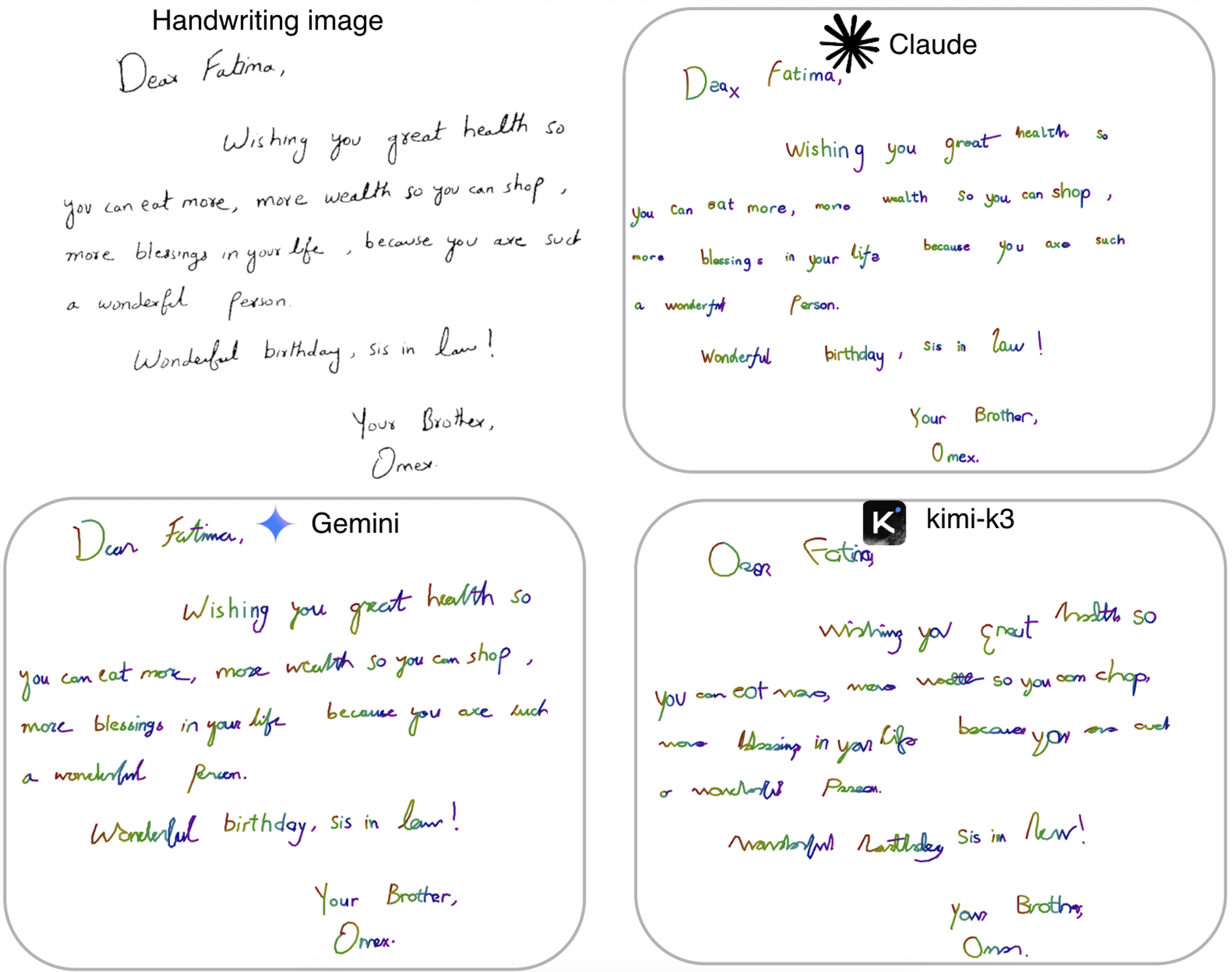}

  \vspace{1em}
  \includegraphics[width=\textwidth,height=0.42\textheight,keepaspectratio]{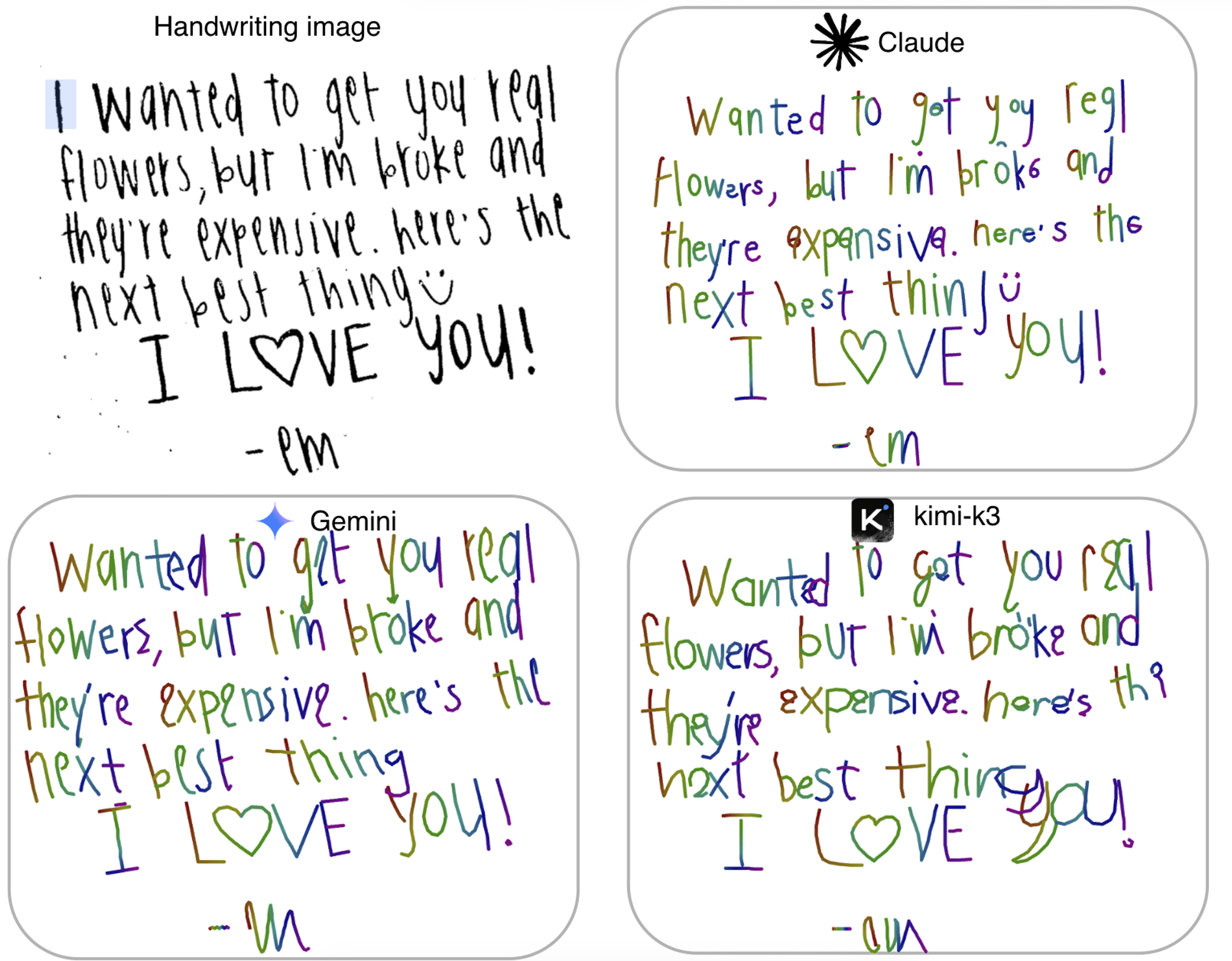}
  \caption{Additional qualitative results (examples 1--2) for Claude-Opus-4.8, Gemini-3.1-Pro and Kimi-K3 on \emph{inkGround}. Top: example 1. Bottom: example 2.}
  \label{figs:qualitative1}
\end{figure*}

\begin{figure*}[p]
  \centering
  \includegraphics[width=\textwidth,height=0.42\textheight,keepaspectratio]{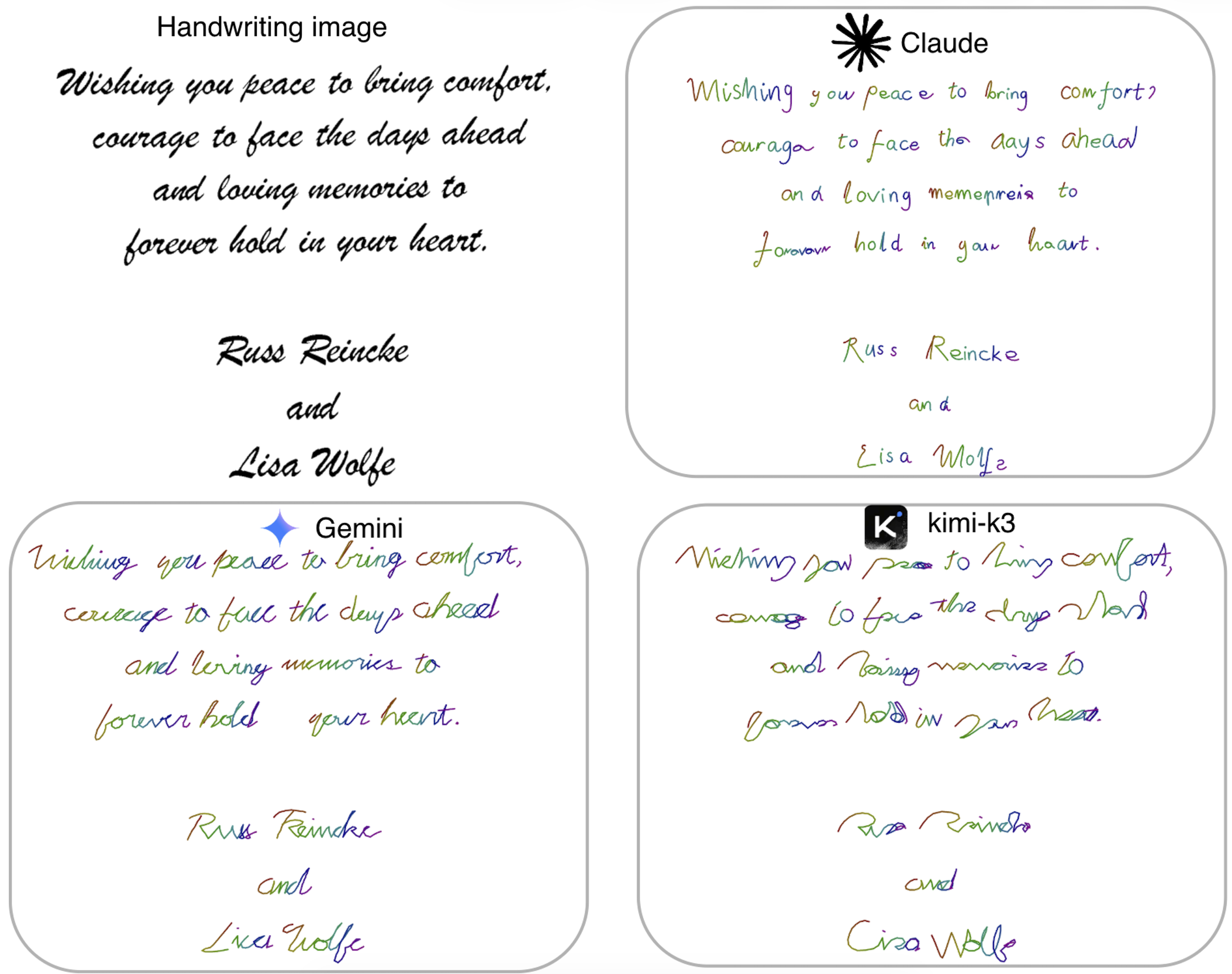}
  \caption{Additional qualitative result (example 3) for Claude-Opus-4.8, Gemini-3.1-Pro and Kimi-K3 on \emph{inkGround}.}
  \label{figs:qualitative2}
\end{figure*}

Figures~\ref{figs:qualitative1} and~\ref{figs:qualitative2} show additional qualitative results from the three closed-source models, in the same layout as Fig.~4 of the main paper.

\clearpage

\subsection*{Appendix E: Data source and acknowledgment}
\label{app:data}
The dataset was provided by a greeting-card company [name withheld for review], which supplied samples of cursive-style handwriting. The company holds the copyright of the dataset and has agreed to its publication for research purposes. We thank the company for its support.

\end{document}